# Forecasting Solar Activity with Two Computational Intelligence Models (A Comparative Study)

M.Parsapoor, U.Bilstrup, B.Svensson

*Abstract*—**Solar activity It is vital to accurately predict solar activity, in order to decrease the plausible damage of electronic equipment in the event of a large high-intensity solar eruption. Recently, we have proposed BELFIS (Brain Emotional Learning-based Fuzzy Inference System) as a tool for the forecasting of chaotic systems. The structure of BELFIS is designed based on the neural structure of fear conditioning. The function of BELFIS is implemented by assigning adaptive networks to the components of the BELFIS structure. This paper especially focuses on performance evaluation of BELFIS as a predictor by forecasting solar cycles 16 to 24. The performance of BELFIS is compared with other computational models used for this purpose, and in particular with adaptive neuro-fuzzy inference system (ANFIS).**

## I. INTRODUCTION

SPACE weather phenomenon, such as solar wind from a coronal mass ejection, can create powerful geomagnetic storms within the Earth's magnetosphere. This sometimes causes harmful effects on electronic and communication systems on earth. For example, geomagnetic storms can interfere or disrupt satellites in low Earth orbit, power grid, radar systems, air traffic control systems, high-frequency communications and global positioning systems (GPS) [1]-[7].

Solar activity is a quasi-periodic space weather phenomenon that causes fluctuating ultraviolet and X-ray emissions. Each solar cycle lasts approximately between 8 to 11 years. According to the National Oceanic and Atmospheric Administration's (NOAA) Space Weather Prediction Center the "monitoring and forecasting solar outbursts in time to reduce its effect on space-based technologies have become a national priority" [2]. The prediction of solar activity includes forecasting different aspects of solar cycles such as solar minima and maxima, cycle period, and cycle amplitude. According to [5], the solar activity forecasting methods aim at predicting "the amplitude of an upcoming solar maximum no later than right after the start of the given solar cycle" [5]. The forecasting methods also have a goal to provide accurate results for long term prediction (several years ahead) of solar activity; however achieving such a long term prediction is hard. As a result, many proposed forecasting methods, including the method given in this paper, are compared based on their performance with much shorter prediction time (few months). A common way to predict solar activity is through the use of statistical-based methods, which encompass a wide variety of methods, such as: linear methods, nonlinear autoregressive methods, linear prediction filter, and computational intelligence models [5], [6]. Moreover, different indices, such as: sunspot numbers, sunspot areas, the 10.7 cm Solar Flux, and geomagnetic activities have been utilized to measure solar activities [1], [3]-[4]. Amongst these, a well-known index is the sunspot numbers, which counts the 'dark regions' ('sunspots') [7] on the sun's surface. The number of sunspots can be calculated according to a formulation called the Wolf Number [7], which estimates the number of sunspots based on the number of sunspot groups and the number of individual sunspots. Computational Intelligence (CI) models (e.g., neural networks (NNs), neuro-fuzzy methods (NFs) and emotion-based data-driven methods [8]-[19]) have also been examined to predict sunspot time series and to forecast its peaks. NNs and NFs have shown reasonable capabilities to capture the chaotic and nonlinear behavior of solar cycles, but they suffer from high time and model complexity [10]. The former depends on the optimization method, the number of training samples, and the dimension of each sample. The latter depends on the number of learning parameters. Note that time and model complexity are worsened when trying to predict future values of a time series with a large number of data samples and high dimensions. It has been observed that in the case when only a small number of data samples are available, the above CI models, suffer from under-fitting issues and fail to predict the long-term activity of chaotic systems such as solar activity [10]. Recently, some techniques have been introduced to increase the prediction accuracy of NNs and NFs in both short-term and long-term perspectives. One of these recent efforts includes the use of pre-processing techniques, e.g., spectral techniques, such as Single Spectrum Analysis (SSA) [8], [10]. Another technique is to modify NFs (e.g., adaptive neuro-fuzzy system (ANFIS)) to improve the capability to perform long-term prediction of chaotic systems accurately. For example, a model that is called the Emotional Learning Fuzzy Inference System (ELFIS) [11], was proposed to decrease the complexity of (ANFIS) and to increase the prediction accuracy of ANFIS by adding an 'emotional cue' [11]. In this paper, we evaluate the performance of the recently proposed Brain Emotional Learning Based Fuzzy Inference System (BELFIS) by examining it as solar cycle predictor for both long term and short term prediction of solar activity. It has been observed that BELFIS, which is a modification of ANFIS, shows better results than ANFIS. BELFIS is based on the neural structure of fear conditioning [20] and the principles of the amygdala-orbitofrontal cortex system [21]. The modifications are described in detail in [18] and [19]. After this introduction, the rest of this paper is organized as follows. Section II briefly illustrates the structural and functional aspects, along with the learning algorithm of BELFIS. Section III evaluates the performance of BELFIS as evidenced in forecasting solar activity. Section IV discusses notable conclusions.

## II. Brain Emotional Learning Based Fuzzy Inference Systems

This section briefly describes how the structure of BELFIS has been designed, how the function of BELFIS has been implemented and how the learning algorithm of BELFIS can be defined.

### A. The structure of BELFIS

The emotional theory that forms the basis of BELFIS is the fear conditioning theory, proposed by LeDoux [20]. The fear conditioning theory describes the neural structure of fear conditioning and emphasizes the role of the amygdala and its internal nuclei such as the lateral (LA) nucleus, basal (B) nucleus and central (CE) nucleus in processing emotional stimuli and providing emotional reactions. Fig. 1 displays the circuits and parts (thalamus, sensory cortex, and orbitofrontal cortex) that are involved in making the association between emotional stimulus and emotional response.

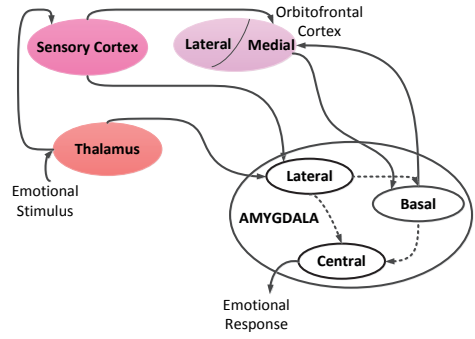

**Fig. 1**. A schematic view of the brain regions that have a role in emotional processing.

The brain regions and circuits in Fig. 1 are the basis of the structure of BELFIS, as depicted in Fig. 2. The model structure includes four main components: TH (Thalamus), CX (sensory CorteX), AMYG (AMYGdala) and ORBI (ORBItofrontal cortex). A detailed description of each part is given in [19].

*B. The Function of BELFIS*

Jang [22] defined adaptive networks that could be represented as a feedforward or a recurrently connected network of adaptive nodes. A good example of adaptive networks is adaptive network neuro-fuzzy inference system (ANFIS). At the training phase, BELFIS receives a pair $\{i, r\}$, provides an output, $r_j$ and adjusts different parameters. The function of the BELFIS is implemented by assigning two types adaptive networks[1], simple adaptive network (SAN) and adaptive neuro fuzzy inference system (ANFIS) to different components of the structure of BELFIS as represented in Fig. 2 (a). As Fig. 2 (b) describes TH is the first part of the BELFIS that receives $i = \{i_1, i_2\}$ and consists of two subparts, MAX_MIN (maximum_minimum) and AGG (aggregation). MAX_MIN consists of two simple adaptive networks (SAN). The output of a MAX_MIN are denoted as $[\max(i), \min(i)]$. The AGG has the role in transforming input vector $i$ to the CX that is responsible to provide $s$ as a vector and send it to AMYG. The AMYG is divided into two parts: the BL (corresponding to the basal and lateral parts of the amygdala) and the CM (corresponding to the accessory basal and cortico-medial region of the amygdala). BL receives $th^{MAX\_MIN}$ and $s$ and provides the primary output $r_a$ as BL sends the primary response, $r_a$ to CM in AMYG. The function of CM can be implemented by assigning an ANFIS that provides $r_j$. RBI also has a connection to AMYG and consists of MO (Medial part of the Orbitofrontal cortex) and LO (Lateral part of the

---

[1] The concept of adaptive networks and different types of adaptive have been defined in [19].

Orbitofrontal cortex). MO receives $s$ and provides the primary output $r_o$ that is sent to CM. The function of BELFIS has been defined based on the function of ANFIS. Moreover, BELFIS is based on ANFIS that has been considered as a universal approximator by proving Stone-Weierstrass theorem. Thus, it can be concluded that BELFIS is also a universal approximator.

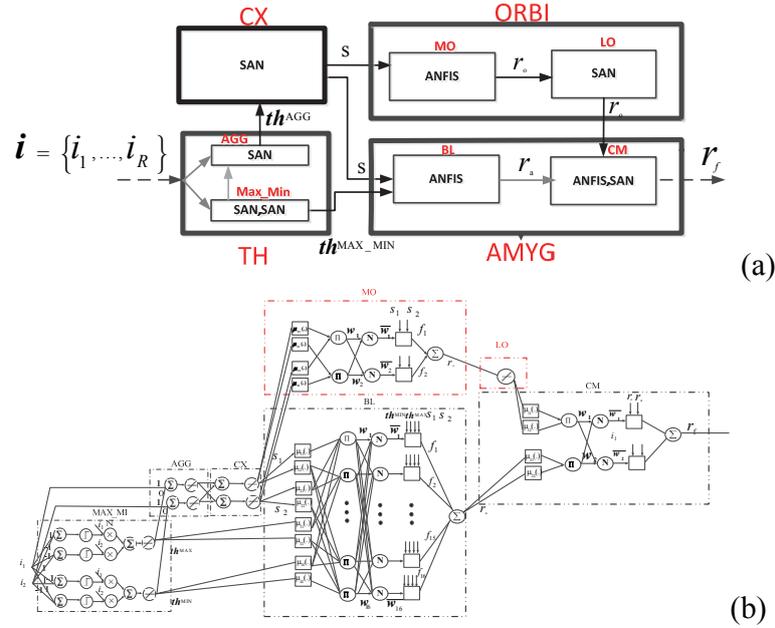

**Fig. 2.** (a). The outlined structure of BELFIS consists of four main parts, TH, CX, ORBI and AMYG. Apart from CX, these parts are divided into two sub-parts. The connections between the different parts are shown as black arrows, while the connections between two sub-parts are presented with grey arrows. (b) A schematic of the connection between different adaptive networks of BELFIS.

*C. The learning aspect of BELFIS*

In general, BELFIS utilizes a hybrid learning algorithm to adjust the parameters. The hybrid learning algorithm is a combination of Steepest Descent (SD) and Least Square Estimator (LSE). To adjust the nonlinear parameters of BL, CM in AMYG and MO in ORBI, an SD based algorithm is used. The LSE is also used to update the consequent parameters and linear learning parameters. The above aspects of BELFIS have been demonstrated in details in [19].

## II. SOLAR ACTIVITY FORECASTING

Various prediction methods, including CI models, have been utilized to forecast the solar activity [8]-

[19]. One well-known indicator of solar activity is the sunspot index, which has been recorded since 1747 [23]. In addition, yearly, monthly and daily sunspot numbers can be downloaded from WDC-SILO and be utilized to reconstruct the SunSpotNumber (SSN) time series [23]. This section assesses the performance of BELFIS by use of two such error measures: normalized mean square error (NMSE) [10] and root mean square error (RMSE) [10]. We examine BELFIS to predict different solar cycles and compare its performance with that of ANFIS. The main goal of this comparison is to show that BELFIS is a powerful method for predicting chaotic systems and they study uses solar cycle 16 to 24 for that purpose. A comparison is also conducted with some NN and NF based methods.

*A. Solar cycles 16, 17, 18*

The first experiment section is related to solar cycles 16, 17 and 18, which have peaks in 1928, 1937 and 1948, respectively. Here, BELFIS and ANFIS are compared by predicting these cycles, and subsequently, the result is also compared with NN and NF based methods. The NN models of this subsection include: WNet (Weight Elimination Feed Forward), a MLP (Multi-Layer Perceptron) with a modified cost function [8], DRNN (Dynamic Recurrent Neural Network) [1] and LogF-NN (gamma Feedback Neural Network) [8] and RBF (Radial Basis Function) [11]. Furthermore, the results obtained from BELFIS are compared with the results of a neuro-fuzzy method, called LLNF (Local linear neuro-fuzzy) [11]. The training data set is chosen from 1700 to 1920 and the test set is chosen from 1920 to 1955. BELFIS and ANFIS are applied to predict one year ahead of yearly SSN time series, which is a time series with yearly numbers of sunspots. It can be seen that BELFIS is more accurate than the majority of the other CI models, as presented in Table I. In the study analysis, the obtained error of BELFIS is close to the obtained error of LLNF, which is the most accurate CI model of Table I. Another interesting point is that the total number of rules in BELFIS is 4 times the number of rules in ANFIS. It might be assumed that increasing the number of rules in ANFIS might decrease the obtained NMSE, however, on attempting to increase the number of rules in ANFIS and in particularly this case, the results have become worse. Figure 3 depicts the predicted values of BELFIS by the grey dashed curve. It is interesting that BELFIS has a reasonable performance to predict the peaks of solar cycles 17 and 18.

TABLE I
COMPARISON OF CI MODELS FOR SOLAR CYCLE 16 TO 18.

| Learning Method | Type of CI model | Specification | NMSE |
|---|---|---|---|
| BELFIS | NF | 16 rules | 0.098 |
| ANFIS | NF | 4 rules | 0.111 |
| LogF-NN[8] | NN | Not identified | 0.112 |
| WNet[8] | NN | Not identified | 0.086 |
| DRNN[8] | NN | Not identified | 0.091 |
| MLP[11] | NN | Not identified | 0.140 |
| RBF[11] | NN | Not identified | 0.118 |
| LLNF[11] | NF | Not identified | 0.070 |

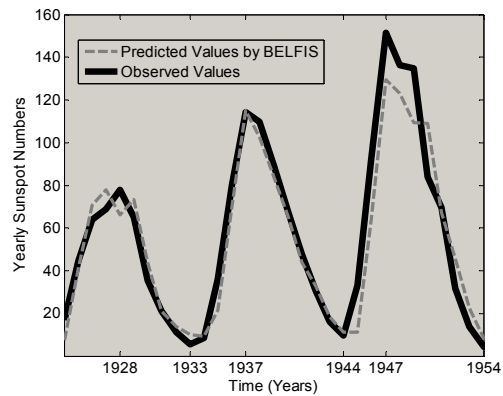

**Fig. 3.** Predicted values of the yearly sunspot number for solar cycles 16-18 using BELFIS (grey dashed curve).

*B. Solar cycle 19*

Solar cycle 19 started in February 1954 and ended in October 1964. The maximum monthly smoothed sunspot number is 201.3. The maximum peak for non-smoothed sunspot number is equal to 253.8 and it occurred in March 1957. In this experiment BELFIS is trained with the use of the non-smoothed monthly sunspot numbers from 1700 to 1950, subsequently, it is tested to predict one month ahead of the monthly SSN time series from 1950 to 1965. This experiment has focused on estimating the peak of solar cycle 19. Table II compares the specification, predicted peak and the NMSE of BELFIS, ANFIS, RBF and ELFIS. It indicates that BELFIS is the most accurate model among the methods. It predicts the peak as 261.25, which is close to the observed peak in 1957. Also, it should be noted that in [11], the predicted peaks of ELFIS and RBF have not been stated and are therefore not included in Table II. Moreover, in this experiment the model complexity of BELFIS is higher in comparison with the other three methods. However, the study shows that BELFIS decomposes the learning parameters for several sets and adjusts each set of the parameters separately, i.e., a divide-and-conquer method to find the optimal values of learning parameters.

TABLE II
COMPARISON OF DIFFERENT METHODS TO PREDICT SOLAR CYCLE 19

| .Learning Method | Specification | Predicted Peaks | NMSE |
|---|---|---|---|
| BELFIS | 16 fuzzy rules | 261.25 | 0.0995 |
| ANFIS | 8 fuzzy rules | 204.91 | 0.1042 |
| ELFIS[11] | 3 fuzzy rules | Not identified | 0.1386 |
| RBF[11] | 7 neurons | Not identified | 0.1314 |

C. *Solar cycles 20, 21 and 22*

Solar cycles' 20, 21 and 22 have peaks in 1968, 1979 and 1989, respectively. This subsection presents the obtained results when BELFIS and ANFIS are applied to predict solar cycles 20, 21 and 22 (the yearly SSN time series from 1965 to 1997). The methods have been trained with the yearly SSN time series from 1700 to 1965. Table III presents NMSE and RMSE from BELFIS, ANFIS and an NN-based model in the form of a MLP with one input layer (six neurons), one hidden layer (12 neurons) and one output layer (one neuron) [28]. Table III also lists the number of rules of BELFIS and ANFIS and the number of neurons of the NN, as well as two error indices, NMSE and RMSE. Additionally, it can be observed that the RMSE obtained from BELFIS is slightly lower than the two other methods. Figure 4 presents how BELFIS predicts these three solar cycles. It is notable that BELFIS is successful in predicting the solar cycles in terms of the time of occurrence of solar maximum. It can be also observed that BELFIS can correctly predict peaks one year ahead for years 1979 and 1989. However, the analysis shows that it is not good at predicting the start of the cycle for years 1977 and 1987.

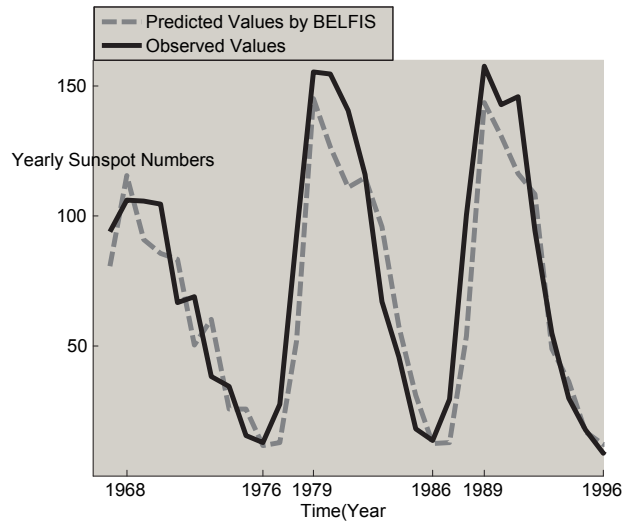

**Fig. 4.** Predicted values of the monthly sunspot number for solar cycle 20-22 using BELFIS (grey dashed lines).

TABLE III
COMPARISON OF DIFFERENT METHODS FOR PREDICTING SOLAR CYCLES 20-22.

| Learning Method | Specification | NMSE | RMSE |
| --- | --- | --- | --- |
| BELFIS | 28 rules | 0.1240 | 18.87 |
| ANFIS | 16 rules | 0.1485 | 19.00 |
| NN [28] | 6-12-1 | Not identified | 22.6 |

*D. Solar cycle 23*

Solar cycle 23 is well-known for being the longest (12.6 years). The starting time of this cycle was May 1996 and it ended in January 2008. The maximum of the smoothed sunspot number occurred in March 2000 and was 120.8. The aim of this experiment is to provide a careful performance comparison between BELFIS and three powerful NNs [24]-[26]. This is achieved by predicting the sunspot numbers one-month ahead. The smoothed monthly sunspot numbers[2] from November 1834 to June 2000 is used, the first 1000 samples are considered as the training data set and the next 1000 samples are used for evaluation. The three NN-based models used for comparison are Hybrid NARX-Elman Recurrent Neural Network [24], the Functional Weights Wavelet Neural Network-based state-dependent Auto-Regressive (FWWNN-AR) model [25] and an evolving neural network (ERNN) [26]. The NARX-Elman model is a hybrid NN that combines a four-layer Elman network with a two-layer NARX neural network. Table IV presents the specification, the NMSE of each model and the embedded dimension of the SSN time series that has been predicted by each method. The most accurate model is NARX-Elman, however, the NMSE performance of BELFIS is close to NARX. It should be noted that in [24], NARX-Elman is combined with a pre-processing method; however, no pre-processing has been used for BELFIS. Also, most likely, using a pre-processing method such as SSA would result in a decreased NMSE for BELFIS too.

---

[2] To calculate the smoothed monthly number, an averaging of monthly mean values over the 13 months is calculated.

*E. Solar cycle 24*

Solar cycle 24, the current cycle, started on January 10$^{th}$ 2008. It is different from other cycles because of the deep solar minimum before its start, its small solar maximum, and its two solar peaks. Hence, the prediction of this solar cycle has captured excessive attention. For this specific cycle, four examples have been used to evaluate BELFIS. In the first example, yearly SSN time series from 1700 to 2008 is used as the training data samples. The trained models (BELFIS and ANFIS) have been then used to recursively predict sunspot numbers from 2009 to 2019. Recursive prediction means that the predicted value in each step is fed to reconstruct the future values of time series sample. Figure 5 depicts the predictions of two models and it is interesting to note that BELFIS and ANFIS have similar performance in predicting solar cycle 24.

In the second experiment, the non-smoothed monthly sunspot numbers from 1700 until April 1996 are used to train the BELFIS and ANFIS. Then, the capability of these models to predict one and three months ahead of SSN time series is investigated. The test data set comprises of non-smoothed monthly sunspot numbers from May 1996 to February 2015, parts of solar cycle 23 and solar cycle 24. As mentioned previously, during solar cycle 24, there are two peaks; for non-smoothed monthly sunspot number the peaks are 96.7 and 102.3. The NMSEs obtained and the predicted values of the first and second peaks of solar cycle 24 are listed in Table V. It shows that BELFIS and ANFIS provide similar results in terms of NMSE.

TABLE IV
COMPARISON OF DIFFERENT METHODS OF PREDICTING SOLAR CYCLE 23 USING SMOOTHED MONTHLY SUNSPOT NUMBERS

| Prediction Model | Specification | NMSE | The embedding |
|---|---|---|---|

|         |              |         | of time |
|---------|--------------|---------|---------|
| BELFIS  | 38           | 7.6e-4  | 4       |
| ANFIS   | 8            | 7.7e-4  | 4       |
| NARX-Elman[24] | 5-7-5-5 3-1 | 5.23e-4 | 5 |
| ERNN[26] | Not identified | 2.8e-3 | Not identified |
| FWWNN[25] | 4-2        | 5.90e-4 | Not identified |

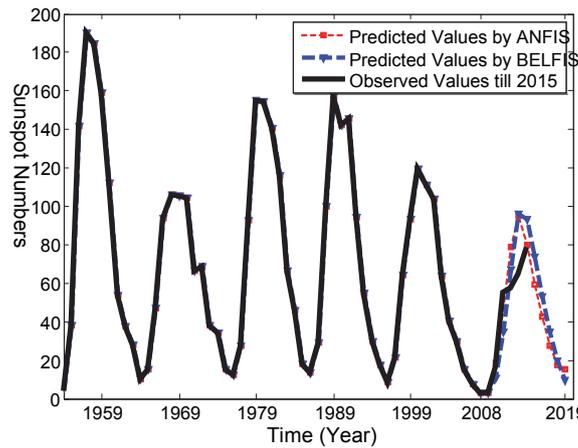

**Fig. 5.** Predicted values for solar cycle 24 using BELFIS and ANFIS.

Figure 6. (a) describes the one month ahead predicted values by BELFIS, while Figure 6. (b) shows the three months ahead predicted values by BELFIS versus the real values of monthly sunspot numbers. In the third example, BELFIS and ANFIS are applied to predict one, five and ten months ahead. The training data and the test data set have been selected from 1700 until May 2007 and June 2007 to September 2014, respectively. The prediction horizon, the predicted second peaks (the observed value of the second peak is 87.6) and the NMSEs of BELFIS and ANFIS are listed in Table VI. Figure 7 shows the change in NMSE of BELFIS and ANFIS with an increase in the horizon of prediction. As can be seen, an increase in the horizon of prediction causes a resultant increase in the NMSE for both methods; however, the NMSEs for BELFIS are less than the NMSEs for ANFIS. The fourth example of this subsection is proposed to compare the performance of BELFIS and two NNs that have been applied to predict solar cycle 24. Table VII presents the training and test data samples, the predicted peaks & time of occurrence and the time of the studies. It can be seen that BELFIS is more successful than the other methods as listed in Table VII.

TABLE V

COMPARISON OF DIFFERENT METHODS FOR PREDICTING SOLAR CYCLE 24 WITH NON-SMOOTHED MONTHLY SUNSPOT NUMBERS

| Learning method | Prediction horizon | Predicted peaks (first, second) | NMSE |
|---|---|---|---|
| BELFIS | One month | 91.51, 92.69 | 0.168 |
| BELFIS | Three months | 79.53, 98.68 | 0.249 |
| ANFIS | One month | 93.32, 95.15 | 0.164 |
| ANFIS | Three months | 79.88, 95.68 | 0.251 |

TABLE VI

COMPARISON OF DIFFERENT METHODS FOR PREDICTING SOLAR CYCLE 24 WITH SMOOTHED MONTHLY SUNSPOT NUMBERS

| Learning method | Prediction horizon (months) | Predicted peaks | NMSE |
|---|---|---|---|
| BELFIS | One | 88.40 | 4.7e-3 |
| BELFIS | Five | 89.48 | 5.2e-2 |
| BELFIS | Ten | 98.61 | 1.8e-1 |
| ANFIS | One | 88.41 | 4.9e-3 |
| ANFIS | Five | 90.74 | 6.7e-2 |
| ANFIS | Ten | 99.7 | 4.6e-1 |

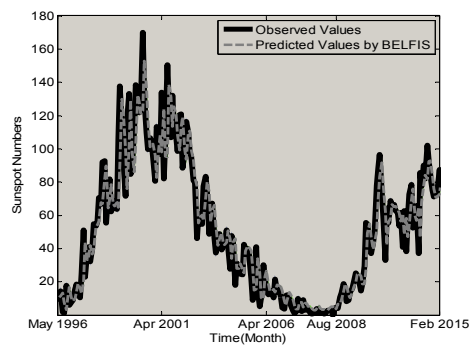

(a)

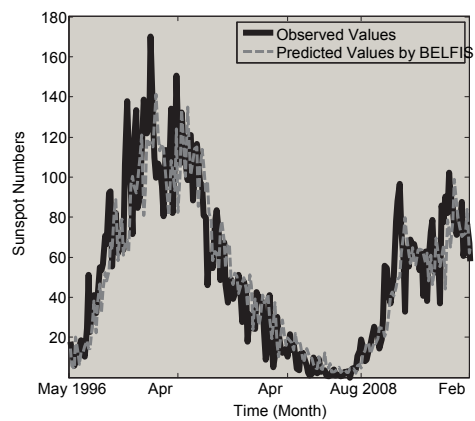

(b)

**Fig. 6.** Predicted non-smoothed monthly SSN time series of solar cycle 24 using BELFIS. (a). Predicted one-month ahead. (b). Predicted three-month ahead.

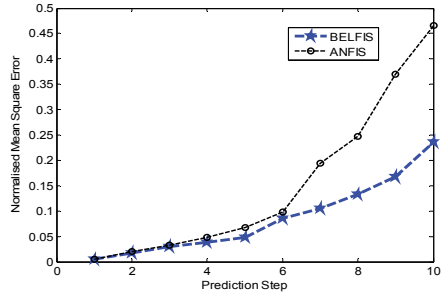

**Fig. 7:** NMSE curves for two methods versus the horizon of prediction.

TABLE VII.
COMPARISON BETWEEN BELFIS AND TWO NNS FOR PREDICTING SOLAR CYCLE 24

| Learning Method | Data Samples | Predicted Time, Peaks | Time, Reference |
|---|---|---|---|
| BELFIS | Monthly SSN Train:1976-2004 Test: 2004-2012 | 2012, 67.13 | 2015, this paper |
| BELFIS | Yearly SSN Train:1882-2009 Test: 2009-2018 | 2013, 67.7 | 2015, this paper |
| Neural Network | Monthly SSN Train:1976-2004 Test: 2004-2012 | 2009, 145 | 2006, [26] |
| Neural network | Yearly SSN Train:1882-2009 Test: 2009-2018 | 2013, 65 | 2011, [27] |

This section presented different examples to extensively examine BELFIS by predicting different solar cycles. Table VIII reports the obtained value for peaks of solar cycles 16-24. It also lists the obtained absolute errors to predict peaks. The maximum error is related to solar cycle 18, and the minimum error is related to solar cycle 16. The results obtained verify that BELFIS has the capability to predict solar activity, as an example of chaotic time series.

TABLE VIII
PRESENTATION OF THE BELFIS'S ABSOLUT ERROR

| Solar Cycle | Prediction horizon | Observed, predicted Peaks | Absolut error |
|---|---|---|---|
| #16 | One year | 77.80, 77.98 | 0.183 |
| #17 | One year | 114.4, 115.2 | 0.798 |
| #18 | One year | 129.4, 151.6 | 37.2 |
| #19 | One month | 253.8, 261.25 | 7.45 |
| #20 | One year | 105.90, 115.62 | 9.72 |
| #21 | One year | 155.40, 145.14 | 10.25 |
| #22 | One year | 157.60, 143.43 | 14.61 |
| #23 | One month | 120.80, 122.48 | 1.68 |
| #24 | One year | 64.90, 96.24 | 31.30 |

### III. CONCLUSION

This paper briefly explained BELFIS and showed how the structure of BELFIS is formed on the basis of the neural structure of fear conditioning. The modular structure of BELFIS is different from other modular CI models in the sense that its structure has been copied from the neural structure of fear conditioning. The learning parameters of BELFIS are divided into separate sets. Each set has its learning algorithms and its learning rules. In general, BELFIS has been developed to improve on ANFIS in chaotic time series prediction. To verify the above claim, we applied BELFIS to predict nine solar cycles (solar cycles 16 to 24) and subsequently, compared its performance with ANFIS and several NN and NF based methods. In all of the examples, BELFIS showed more accurate results than ANFIS. In addition, on comparison with other NNs and NF methods, BELFIS provided reasonable performance, which verifies that BELFIS is a valid CI model for space weather prediction. BELFIS aims to address the model and time complexity issues. The good results of BELFIS in predicting non-smoothed sunspot numbers indicate that BELFIS is a model with low noise sensitivity.